\pdfoutput=1

\documentclass[11pt]{article}

\usepackage[]{acl}

\usepackage{times}
\usepackage{latexsym}

\usepackage[T1]{fontenc}

\usepackage[utf8]{inputenc}

\usepackage{microtype}

\usepackage[linesnumbered, ruled]{algorithm2e}
\usepackage[T1]{fontenc}
\usepackage{dsfont}

\usepackage{mdframed}
\usepackage{booktabs}
\usepackage{graphicx}
\usepackage{subcaption}

\usepackage[normalem]{ulem}
\usepackage{enumitem}
\usepackage{xcolor}
\usepackage{soul}
\colorlet{soulred}{red!30}
\sethlcolor{soulred}%

\usepackage{amsmath,amsthm,amsfonts,amssymb,bm}
\DeclareMathOperator*{\argmax}{arg\,max} 
\usepackage{framed}
\usepackage{varioref}
\usepackage{float}
\usepackage{multirow}
\usepackage{dashrule}
\usepackage{url}
\usepackage{pifont}
\newcommand{\eat}[1]{\ignorespaces}

\usepackage{xparse}
\NewDocumentCommand{\heng}
{ mO{} }{\textcolor{red}{\textsuperscript{\textit{Heng}}\textsf{\textbf{\small[#1]}}}}

\title{Dynamic Global Memory for Document-level Argument Extraction}

\author{Xinya Du \quad \quad Sha Li \quad \quad Heng Ji \\
Department of Computer Science\\
University of Illinois Urbana-Champaign \\ 
{\tt \{xinyadu2,shal2,hengji\}@illinois.edu}\\
}

\begin{document}
\maketitle
\begin{abstract}
Extracting informative arguments of events from news articles is a challenging problem in information extraction, which requires a global contextual understanding of each document. While recent work on document-level extraction has gone beyond single-sentence and increased the cross-sentence inference capability of end-to-end models, they are still restricted by certain input sequence length constraints and usually ignore the global context between events.
To tackle this issue, we introduce a new global neural generation-based framework for document-level event argument extraction by constructing a document memory store to record the contextual event information and leveraging it to implicitly and explicitly help with decoding of arguments for later events.
Empirical results show that our framework outperforms prior methods substantially and it is more robust to adversarially annotated examples with our constrained decoding design.\footnote{Our code and resources are available at \url{https://github.com/xinyadu/memory_docie} for research purpose.}
\end{abstract}

\section{Introduction}

An event is a specific occurrence involving participants (people, objects, etc.). Understanding events in the text is necessary for building machine reading systems, as well as for downstream tasks such as information retrieval, knowledge base population, and trend analysis of real-life world events~\cite{sundheim-1992-overview}.
Event Extraction has long been studied as a local sentence-level task~\cite{grishman-sundheim-1996-message, ji2008refining, grishman_2019,LinACL2020}. This has driven researchers to focus on developing approaches for sentence-level predicate-argument extraction. This is problematic when events and their arguments spread across multiple sentences -- in real-world cases, events are often written throughout a document.\footnote{In {\sc WikiEvents}~\cite{li-etal-2021-document}, nearly 40\% of events have an argument outside the sentence containing the trigger.}

\begin{figure}[t]
\centering
\resizebox{\columnwidth}{!}{
\includegraphics{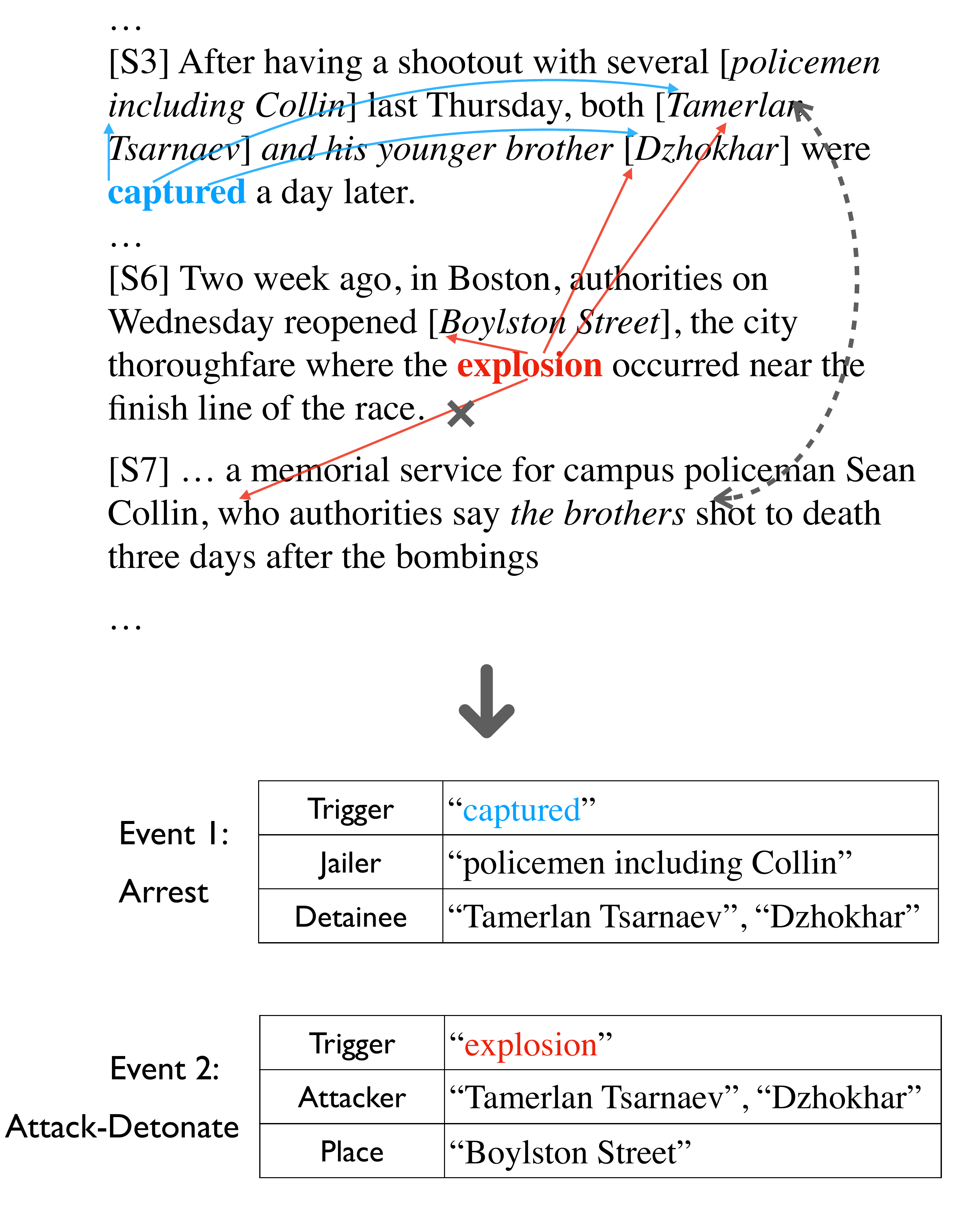}
}
\caption{Document-level event argument extraction.}
\label{fig:task}
\end{figure}

In Figure~\ref{fig:task}, the excerpt of a news article describes two events in the 3rd sentence (an arrest event triggered by ``captured'') and the 6th sentence (an attack event triggered by ``explosion''). 
S6 on its own contains little information about the arguments/participants of the explosion event, but together with the context of S3 and S7, we can find the informative arguments for the \textsc{attacker} role. In this work, we focus on the \textit{informative argument} extraction problem, which is more practical and requires much a broader view of cross-sentence context~\cite{li-etal-2021-document}.
 For example, although ``the brothers'' also refers to ``Tamerlan T.'' and ``Dzhokhar'' (and closer to the trigger word), it should not be extracted as an informative argument.

In recent years, there have been efforts focusing on event extraction beyond sentence boundaries with end-to-end learning~\cite{ebner-etal-2020-multi, du-thesis-2021, li-etal-2021-document}. Most of the work still focuses on modeling each event independently~\cite{li-etal-2021-document} and ignores the global context partially because of the pretrained models' length limit and their lack of attention for distant context~\cite{khandelwal-etal-2018-sharp}. 
\newcite{du-etal-2021-template} propose to model dependency between events directly via the design of generation output format, yet it is not able to handle longer documents with more events -- whereas in real-world news articles there are often more than fifteen inter-related events (Table~\ref{tab:datastats}). 

In addition, previous work often overlooks the consistency between extracted event structures across the long document. For example, if one person has been identified as a \textsc{jailer} in an event, it's unlikely that the same person is an \textsc{attacker} in another event in the document (Figure~\ref{fig:task}), according to world event knowledge~\cite{Sap-etal-2019-atomic, yao-etal-2020-weakly}.

\begin{figure*}[t]
\centering
\resizebox{\textwidth}{!}{
\includegraphics{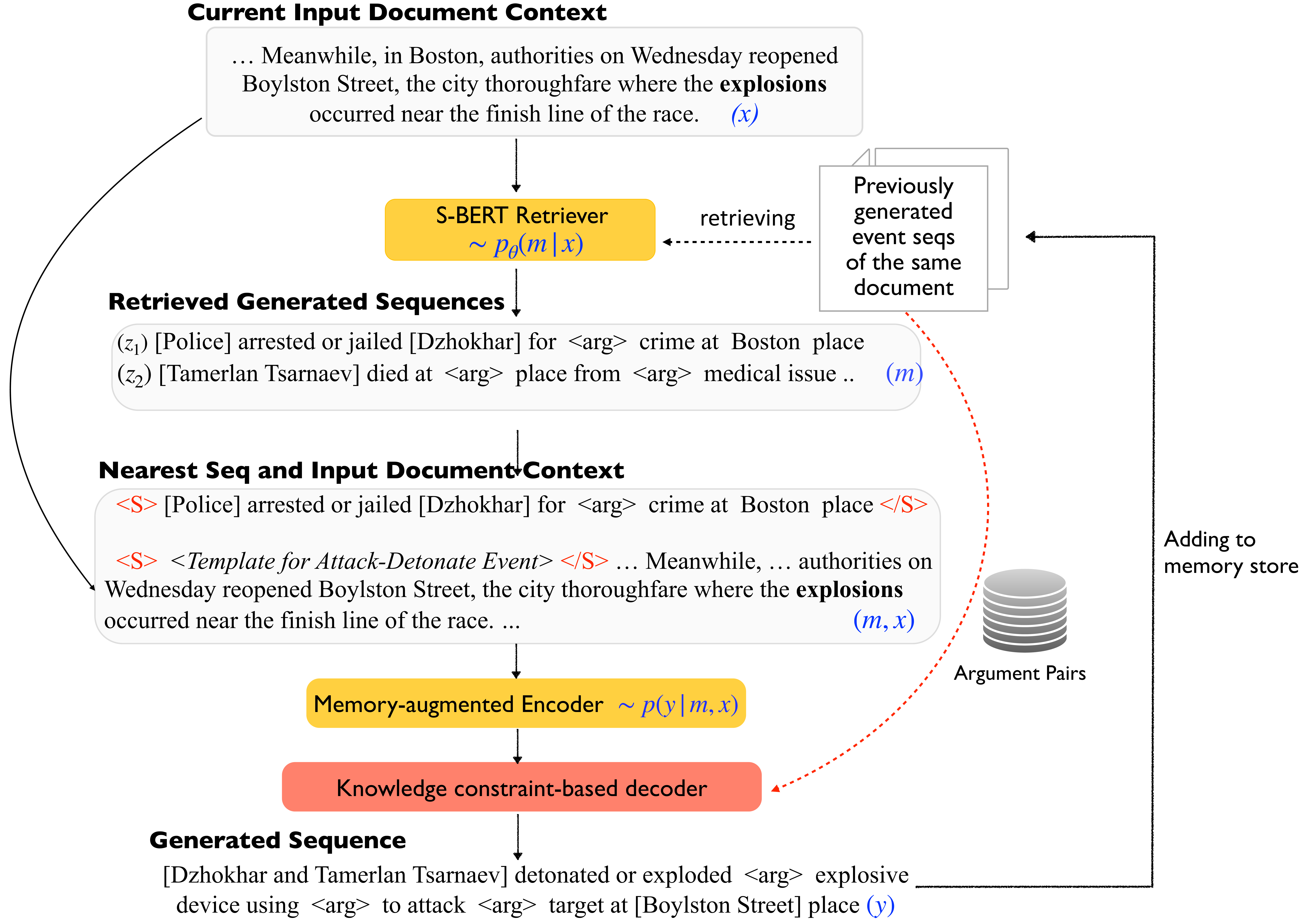}
}
\caption{Our Framework for Memory-enhanced Training and Decoding.}
\label{fig:framework}
\end{figure*}

In this paper, to tackle these challenges and have more consistent/coherent extraction results, we propose a document-level memory-enhanced training and decoding framework (Figure~\ref{fig:framework}) for the problem. It can leverage relevant and necessary context beyond the length constraint of end-to-end models, by using the idea of a dynamic memory store. It helps the model leverage previously generated/extracted event information during both training (implicitly) and during test/decoding (explicitly). 
More specifically, during training, it retrieves the most similar event sequence in the memory store as additional input context to mode. Plus, it performs constrained decoding based on the memory store and our harvested global knowledge-based argument pairs from the ontology.

We conduct extensive experiments and analysis on the {\sc WikiEvents} corpus and show that our framework significantly outperforms previous methods either based on neural sequence labeling or text generation. We also demonstrate that the framework achieves larger gains over baseline non memory-based models as the number of events grows in the document, and it is more robust to manually designed adversarial examples.



\section{Task Definition}
In this work, we focus on the challenging problem of extracting {\bf informative arguments of events}\footnote{Name entity mentions are recognized as more informative than nominal mentions.} from the document.
Each event consists of (1) a trigger expression which is a continuous span in the document, it is of a type $E$ which is predefined in an ontology; (2) and a set of arguments $\{arg_1, arg_2, ...\}$, each of them has a role predefined in the ontology, for event type $E$. In the annotation guideline/ontology, the ``template'' that describes the connections between arguments of the event type is also provided.
For example, when $E$ is {\it Arrest}, its corresponding arguments to be extracted should have roles: \textsc{Jailer} (\verb|<arg1>|), {\sc Detainee} (\verb|<arg2>|), {\sc Crime} (\verb|<arg3>|), {\sc Place} (\verb|<arg4>|). Its description template is:

\begin{quote}
\verb|<arg1>| arrested or jailed \verb|<arg2>| for \verb|<arg3>| crime at \verb|<arg4>| place
\end{quote}
Given a long news document $Doc = \{..., \text{<Trg1>}, ..., x_i, ..., \text{<Trg2>},..., x_n\}$ with given event triggers,
our goal is to extract all the informative argument spans to fill in the role of $E1$, $E2$, etc. For the example piece in Figure~\ref{fig:task}, $E1$ is {\it Arrest} (triggered by <Trg1> ``captured'') and $E2$ is {\it Attack-Detonate} (<Trg2> is ``explosion'').

The ontology is constructed by the DARPA KAIROS project\footnote{\footnotesize https://www.darpa.mil/news-events/2019-01-04} 
for event annotation. It defines 67 event types in a three-level hierarchy, which is richer than the ACE05 ontology with only 33 event types for sentence-level extraction.

\section{Methodology}

In this section, we describe our memory-enhanced neural generation-based framework (Figure~\ref{fig:framework}) for extracting informative event arguments from the document.
%
Our base model is based on a sequence-to-sequence pretrained language model for text generation. We first introduce how we leverage previously extracted events as additional context for training the text generation-based event extraction model to help the model automatically capture event dependency knowledge (Section~\ref{subsec:memorytrain}). To {\it explicitly} help the model satisfy the global event knowledge-based constraints (e.g., it is improbable that one person would be {\sc Jailer} in event A and then {\sc Attacker} in event B), we propose a dynamic decoding process with world knowledge-based argument pair constraints (Section~\ref{subsec:dynamicde}).

\subsection{Memory-enhanced Generation Model \\ for Argument Extraction}
\label{subsec:memorytrain}

Following \newcite{li-etal-2021-document}, the main model of our framework is based on the pretrained encoder-decoder model BART~\cite{lewis-etal-2020-bart}. The intuition behind using BART for the extraction task is that it is pre-trained as a denoising autoencoder -- reconstruct the original input sequence. This fits our objective of extracting argument spans from the input document because the extracted arguments' tokens are from the input sequence.
The generation model takes (1) {\it context:} the concatenation of the piece of text $x$ (of document $D$) containing the current event trigger\footnote{Up to the maximum length limit of the pre-trained model.} and the event type's corresponding template in the ontology; (2) {\it memory store} $m$: of previously extracted events of the same document $D$, as input, and learns a distribution $p(y|x,m)$ over possible outputs $y$. 
The ground truth sequence $y$ is a sequence of a template where the placeholder $\verb|<arg>|$s are filled by the gold-standard  argument spans of the current event.\footnote{The gold sequence for the 1st event in Figure~\ref{fig:task} would be ``[policemen including Collin] arrested or jailed [Tamerlan T. and Dzhokhar] for <arg> crime at <arg> place''}

\begin{equation}
p(y|x,m) = \prod_{i}^{N} p_(y_i|y_{1:i-1},x,m)
\end{equation}

To be more specific on building the dependency between events across the document, we use the most relevant event in the memory store $m$ as additional context, instead of the entire memory store. To retrieve the most relevant ``event'' (i.e., a generated sequence) from the memory store $m = \{m_1, m_2,... \}$, we use S-BERT~\cite{reimers-gurevych-2019-sentence} for dense retrieval (i.e., retrieval with dense representations provided by NN). S-BERT is a modification of the BERT model~\cite{devlin-etal-2019-bert} that uses siamese and triplet network structures to obtain semantically meaningful embeddings for text sequences. We can compare the distance between two input sequences with cosine-similarity in an easier and faster way.
Given a current input document piece $x$, we encode all of the previously generated event sequences in the memory store and $x$. Then we calculate the similarity scores with vector space cosine-similarity and normalization: 

\begin{equation}
\nonumber
\begin{gathered}
    \text{score}(m_i|x) = \frac{\exp{f(x,m_i)}}{\sum_{m_i \in m}{\exp{f(x,m_i)}}} \\
    f(x,m_i) = Embed(x)^T Embed(m_i)
\end{gathered}
\end{equation}
Afterwards, we select the $m_i$ with the highest similarity score: $m^{R} = \argmax_i \text{score}(m_i|x)$

To summarize, the input sequence for the memory-enhanced model consists of the retrieved generated event sequence ($m^{R}$), the template for the current event type ($T$) -- provided by the ontology/dataset, and the context words from the document ($x_1$, ..., $x_n$):

\begin{equation}
\begin{gathered}
\nonumber
\text{<S>} \ m^{R}_1, m^{R}_2, ...,  \ \text{</S>}  \\
\text{<S>} \ T_1, T_2, ... \ \text{</S>} \quad  x_1, x_2, ..., x_n \ \text{[EOS]} \\
\end{gathered}
\end{equation}

During training time, the memory store consists of gold-standard event sequences -- while at test time, it contains real generated event sequences. The training objective is to minimize the negative log likelihood over all $((x, m^{R}, T), y)$ instances.
Since we fix the parameters from S-BERT, the retrieval module's parameters are not updated during training.
Thus the training time cost of our memory-based training is almost the same to the simple generation-based model.

\subsection{Constrained Decoding with \\ Global Knowledge-based Argument Pairs}
\label{subsec:dynamicde}

The constrained/dynamic decoding is an important stage in our framework.
We first harvest a number of world knowledge-based event argument pairs that are probable/improbable of happening with the same entity being the argument. For example, (<Event Type: Arrest, Argument Role: {\sc Jailer}> | <Event Type: Attack-Detonate, Argument Role: {\sc Attacker}>) is an improbable pair. In the framework (Figure~\ref{fig:framework}), they are called ``argument pairs''.
Then based on the argument pairs constraints, the dynamic decoding is conducted throughout the document -- if one entity is decoded in an event in the earlier part of the document, it should not be decoded later in another event if the results are incompatible with the improbable argument pairs.

\begin{algorithm}[t]
\small
\SetKwInOut{Input}{Input}\SetKwInOut{Output}{Output}
  \Input{
        Event Ontology $O$, consisting of $|O|$ events' information. For each event $E_i \in O$, it has a set of argument roles $(A^{i}_1, A^{i}_2,...)$. 
        }
  \Output{A set of (<Event Type, Argument Role> | <Event Type, Argument Role>) pairs with ``probable'' or ``improbable'' denotation.}
  \BlankLine
  $impro\_arg\_pairs \longleftarrow \{\}$\;
  $pro\_arg\_pairs \longleftarrow \{\}$\;
  \tcp{Enumerate event type pairs}
  \For{$i \leftarrow 1$ \KwTo $|O|$}{
    \For{$j \leftarrow i+1$ \KwTo $|O|$}{
        $cnt(i,j)$ = count \# of $(E_i, E_j)$ co-occurrence in the training documents;
        
        \lIf{$\text{cnt}(i,j) == 0$}{continue} 
        \tcp{Enumerate argument pairs}
        \For{$A^{i}_k \in E_i$ args $(A^{i}_1, A^{i}_2,...)$}{
            \For{$A^{j}_h \in E_j$ args $(A^{j}_1, A^{j}_2,...)$}{
                
                \lIf{$entity\_type(A^{i}_k) != entity\_type(A^{j}_h$)}{continue} 
                
                $cnt\_args(A^{i}_k, A^{j}_h)$ = count \# of $(A^{i}_k, A^{j}_h)$ being the same entity in the training set documents;
                
                \lIf{$\frac{cnt\_args(A^{i}_k, A^{j}_h)}{cnt(i,j)}>0.001$}
                {$impro\_arg\_pairs$.add($(<E_i, A^{i}_k> | <E_j, A^{j}_h>)$)}
                \Else{$pro\_arg\_pairs$.add($(<E_i, A^{i}_k> | <E_j, A^{j}_h>)$)}
            }
        }
    }
  }
\caption{\small Automatic Harvesting Argument Pairs from the Event Ontology}
 \label{algo:harvest}
\end{algorithm}

\paragraph{Harvesting Global Knowledge-based Argument Pairs from the Ontology}
%
We first run an algorithm to automatically harvest all candidate argument pairs (Algorithm~\ref{algo:harvest}). Basically, we
\begin{itemize}[leftmargin=*]
    \item First enumerate all possible event type pairs, and count how many times they co-occur in the training set (Line 2--6).
    \item Then we enumerate all possible argument types pairs that share the same entity type from the ontology (e.g., argument {\sc Organization} (ORG) and argument {\sc Victim} (PER) don't have the same entity type),
and count how many times both of the args are of the same entity in training docs (e.g., ``Dzhokhar'' are both {\sc Detainee} and {\sc Attacker} in two events in Figure~\ref{fig:task}) (Line 7--11).
    \item Finally we add into the set of probable argument pairs, whose normalized score is above a threshold (99\% of the candidate arguments with non-zero score);
    and the rest into the set of improbable pairs (Line 11--14).
\end{itemize}

\begin{figure}[t]
\centering
\resizebox{\columnwidth}{!}{
\includegraphics{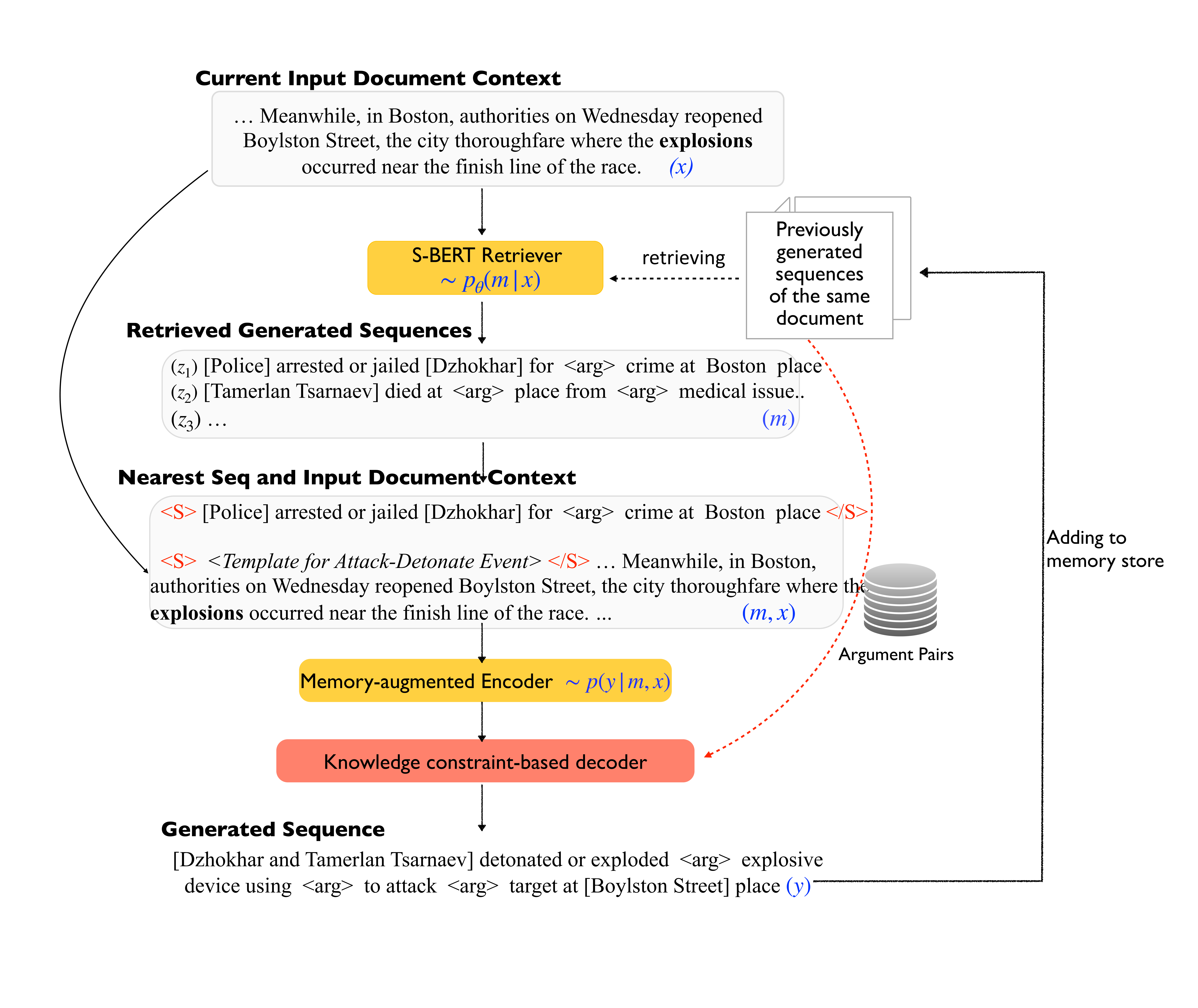}
}
\caption{Constrained/Dynamic Decoding.}
\label{fig:decoding}
\end{figure}

After automatic harvesting, since there is noise in the dataset as well as cases not covered, we conduct a human curation process to mark certain improbable argument pairs as probable, based on world knowledge. Finally, we obtain 1,568 improbable argument pairs and 687 probable pairs.

\begin{table}[!h]
\small \centering
\begin{tabular}{l|cc}
\toprule
           & \begin{tabular}[c]{@{}l@{}}\# pairs with global \\  co-occurrence stats\end{tabular} & \begin{tabular}[c]{@{}l@{}}\# pairs after \\ human curation\end{tabular} \\ \midrule
improbable   & 1,855                                    & 1,568                           \\
probable & 400                                     & 687                          \\ \bottomrule
\end{tabular}
\caption{Statistics of Harvested Argument Pairs.}
\label{tab:harveststats}
\end{table}

\paragraph{Dynamic Decoding Process}

During the decoding process, we keep an explicit data structure in the memory store, to record what entities have been decoded and what argument roles they are assigned to (Figure~\ref{fig:decoding}). During decoding the arguments of later events in the document, assuming we are at a time step $t$ for generating the sequence for event $E_i$, to generate token $y_t$, we first determine the argument role ($A_k$) it corresponds to. Then we search through the memory store if there are extracted entities $e$ that have argument role $A_h$, where $<A_k, A_h>$ is an improbable argument pair. Then when decoding to token at time step $t$, we decrease the probability (after softmax) of generating/extracting tokens in entity $e$ according to the improbable argument pair rule. 
Compared to decreasing the probability of extracting certain conflicting entities, we are more reserved in utilizing the probable argument pairs, only if the same entity has been assigned the argument role for more than 5 times in the document, we are increasing the probability of extracting the same entity (generating the token of the entity) for the corresponding argument role (the most co-occurred).

After the generation process for the current event, we add the newly generated event sequence (extracted arguments) back into the memory store.

\section{Experiments}

\subsection{Dataset and Evaluation Metrics}
We conduct evaluations on the newly released \textsc{WikiEvents} dataset~\cite{li-etal-2021-document}. As compared to the ACE05\footnote{ http://www.itl.nist.gov/iad/mig/tests/ace/2005/} sentence-level extraction benchmark, {\sc WikiEvents} focuses on  annotations for informative arguments and for multiple events in the document-level event extraction setting, and is the only benchmark dataset for this purpose to now. It contains real-world news articles annotated with the DARPA KAIROS ontology. 
%
As shown in the dataset paper, the distance between informative arguments and event trigger is 10 times larger than the distance between local/uninformative arguments
(including pronouns) and event triggers. This demonstrates more needs for modeling long document context and event dependency and thus it requires a good benchmark for evaluating our proposed models. The statistics of the dataset are shown in Table~\ref{tab:datastats}. We use the same data split and preprocessing step as in the previous work.
\begin{table}[h]
\small \centering
\begin{tabular}{l|ccc}
\toprule
                  & Train & Dev & Test \\ \midrule
Documents           & 206   & 20  & 20   \\
Sentences      & 5262  & 378 & 492  \\
Avg. number of events & 15.73 &	17.25 &	18.25 \\
Avg. number of tokens & 789.33 & 643.75	& 712.00 \\ \bottomrule
\end{tabular}
\caption{Dataset Statistics}
\label{tab:datastats}
\end{table}

\begin{table*}[t]
\centering
\resizebox{\textwidth}{!}{
\begin{tabular}{l|cccccc|cccccc} \toprule
 & \multicolumn{6}{c|}{Argument Identification}  & \multicolumn{6}{c}{Argument Classification}                       \\
Models & \multicolumn{3}{c}{Head Match} & \multicolumn{3}{l|}{Coref Match} & \multicolumn{3}{c}{Head Match} & \multicolumn{3}{l}{Coref Match} \\ 
    & P        & R        & F1       & P         & R        & F1       & P        & R        & F1       & P         & R        & F1       \\ \midrule
\begin{tabular}[c]{@{}l@{}}BERT-CRF \\ \cite{shi-lin-2019-BERTCRF}\end{tabular}           & -        & -        & 52.71    & -         & -        & 58.12    & -        & -        & 43.29    & -         & -        & 47.70    \\
\begin{tabular}[c]{@{}l@{}}BART-Gen \\ \cite{li-etal-2021-document}\end{tabular}             & 58.62    & 55.64    & 57.09    & 62.84     & 59.64    & 61.19    & 54.02    & 51.27    & 52.61    & 57.47     & 54.55    & 55.97    \\ \midrule
Memory-based Training          & 61.07    & 56.18    & 58.52    & 66.21     & 60.91    & 63.45    & 55.93    & 51.45    & 53.60    & 60.47     & 55.64    & 57.95    \\
\begin{tabular}[c]{@{}l@{}}\quad w/ knowledge \\ \quad constrained decoding\end{tabular} & 62.45 & 56.55 & {\bf 59.35}     & 67.67 & 61.27 & {\bf 64.31}$^{*}$    & 57.23 & 51.82 & {\bf 54.39} & 61.85 & 56.00 & {\bf 58.78}$^{*}$    \\ \bottomrule
\end{tabular}}
\caption{Performance (\%) on the informative argument extraction task. $^{*}$ indicates statistical significance ($p<0.05$).}
\label{tab:performance-general}
\end{table*}

As for evaluation, we use the same criteria as in previous work. We consider an argument span to be correctly identified if its offsets match any of the gold/reference informative arguments of the current event (i.e., argument identification); and it is correctly classified if its semantic role also matches (i.e., argument classification)~\cite{li-etal-2013-joint}.

To judge whether the extracted argument and the gold-standard argument span match, since the exact match is too strict that  some correct candidates are considered as spurious (e.g., ``the 22 policemen'' and ``22 policemen'' do not match under the exact match standard). Following~\newcite{huang2012modeling, li-etal-2021-document}, we use head word match F1 ({\it Head F1}).
We also report performance under a more lenient metric ``{\it Coref F1}'': the extracted argument span gets full credit if it is coreferential with the gold-standard arguments~\cite{ji-grishman-2008-refining}. The coreference links information between informative arguments across the document are given in the gold annotations.

\subsection{Results}

We compare our framework to a number of competitive baselines. \cite{shi-lin-2019-BERTCRF} is a popular baseline for semantic role labeling (predicate-argument prediction). It performs sequence labeling based on automatically extracted features from BERT~\cite{devlin-etal-2019-bert} and uses Conditional Random Fields~\cite{DBLP:conf/icml/LaffertyMP01} for structured prediction ({\bf BERT-CRF}). \newcite{li-etal-2021-document} propose to use conditional neural text generation model for the document-level argument extraction problem, it handles each event in isolation ({\bf BART-Gen}).
%

For our proposed memory-enhanced training with retrieved additional context, we denote it as {\bf Memory-based Training}. We also present the argument pairs constrained decoding results separately to see both components' contributions.\footnote{All significance tests for F-1 are computed using the paired bootstrap procedure of 5k samples of generated sequences~\cite{berg-etal-2012-empirical}.} 

In Table~\ref{tab:performance-general}, we present the main results for the document-level informative argument extraction. The score for argument identification is strictly higher than argument classification since it only requires span offset match. We observe that:
\begin{itemize}

    \item The neural generation-based models (BART-Gen and our framework) are superior in this document-level informative argument extraction problem, as compared to the sequence labeling-based approaches. Plus, generation-based methods only require one pass as compared to span enumeration-based methods~\cite{wadden-etal-2019-entity, du-cardie-2020-event}.
    \item As compared to the raw BART-Gen, with our memory-based training -- leveraging previously closest extracted event information substantially helps increase precision (P) and F-1 scores, with smaller but notable improvement in recall especially under Coref Match. 
    \item With additional argument pair constrained decoding, there is an additional significant improvement in precision and F-1 scores. This can be mainly attributed to two factors: (I) during constrained decoding, we relied more on ``improbable arg. pairs'' as a checklist to make sure that the same entity not generated for conflicting argument roles in the same document, and only utilize very few top ``probable arg. pairs'' for promoting the decoding for frequently appearing entities; (II) If an entity has been decoded in previous event A by mistake then under the argument pair rule, it will not be decoded in event B even if it correct -- which might hurt the recall. 
\end{itemize}

\begin{table}[t]
\small \centering
\begin{tabular}{l|cccc}
\toprule
     &	\multicolumn{2}{c}{Arg. Classification}  \\
      & Head M.            & Coref. M. \\ \midrule
BART-Gen     & 50.00 &  53.12 \\ 
Memory-based Training & 50.75 & 53.73 \\
\begin{tabular}[c]{@{}l@{}}Our Best Model (w/ knowledge \\ constrained decoding)\end{tabular} & 53.73 & 56.72 \\ \bottomrule
\end{tabular}
\caption{Performance (\%) on adversarial examples.}
\label{tab:adv}
\end{table}

\paragraph{Robustness to Adversarial Examples}
To test how the models react to specially designed adversarial examples, we select a quarter of documents from the original test set, and add one more adversarial event into each of them by adding a few new sentences. 
The additional event is designed to ``attract'' the model to make mistakes that are against our global knowledge-based argument pair rules.\footnote{In our open-sourced repository, readers will be able to find our designed adversarial examples under the data folder.}
An excerpt for one example:
\begin{quote}
    Tandy, then 19, {\bf talks} to his close friend, Stephen Silva, about ... 
    Tandy and Silva both {\bf died} as lifeguards together at the Harvard pool.
    Later a kid was {\bf killed} by a Stephen Silva-lookalike guy.
\end{quote}
In this example, we know ``Stephen Silva'' died in the second event ``Life.Die'' triggered by {\bf died}. Although it is also mentioned in the last sentence, ``Stephen Silva'' should not be extracted as the {\sc Killer}.
In Table~\ref{tab:adv}, we summarize the F-1 scores of argument classification models. Firstly we see on the adversarial examples, the performance scores all drop as compared to the normal setting (Table~\ref{tab:performance-general}), proving it's harder to maintain robustness in this setting. Our best model with argument pair constrained decoding outperforms substantially both BART-Gen and our memory-based training model. The gap is larger than the general evaluation setting, which shows the advantage of {\it explicitly} enforcing the reasoning/constraint rules.

\section{Further Analysis}

In this section, we further provide more insights with quantitative and qualitative analysis, as well as error analysis for the remaining challenges.

\paragraph{Influence of Similarity-based Retrieval}
In Table~\ref{tab:ablation}, we first investigate what happens when our similarity-based retrieval module is removed --  we find that the F-1 scores substantially drop. There's also a drop of scores across metrics when we retrieve a random event from the memory store. It is interesting that the model gets slightly better performance with random memory than not using any retrieved/demonstration sequences. This corresponds to the findings in other domains of NLP on how demonstrations lead to performance gain when using pre-trained language models (especially in the few-shot learning setting).

\begin{table}[h]
\centering
\resizebox{\columnwidth}{!}{
\begin{tabular}{l|cc|cc}
\toprule
     \multirow{2}{*}{Models}              & \multicolumn{2}{c|}{Arg. I.} & \multicolumn{2}{c}{Arg. C.} \\ 
      & H. M.            & C. M.           & H. M.            & C. M.    \\ \midrule
Memory-based Training & 58.52 & 63.45 & 53.60 & 57.95 \\
\quad w/o retrieval      & 56.84 & 61.82 & 51.29 & 55.69 \\ 
\quad w/  random memory	 & 57.65 & 62.69 & 52.22 & 57.17 \\ \bottomrule
\end{tabular}
}
\caption{Ablation (\%) for similarity-based retrieval.}
\label{tab:ablation}
\end{table}

\begin{figure}[h]
    \begin{subfigure}[b]{\columnwidth}
    \centering
    \includegraphics[width=\columnwidth]{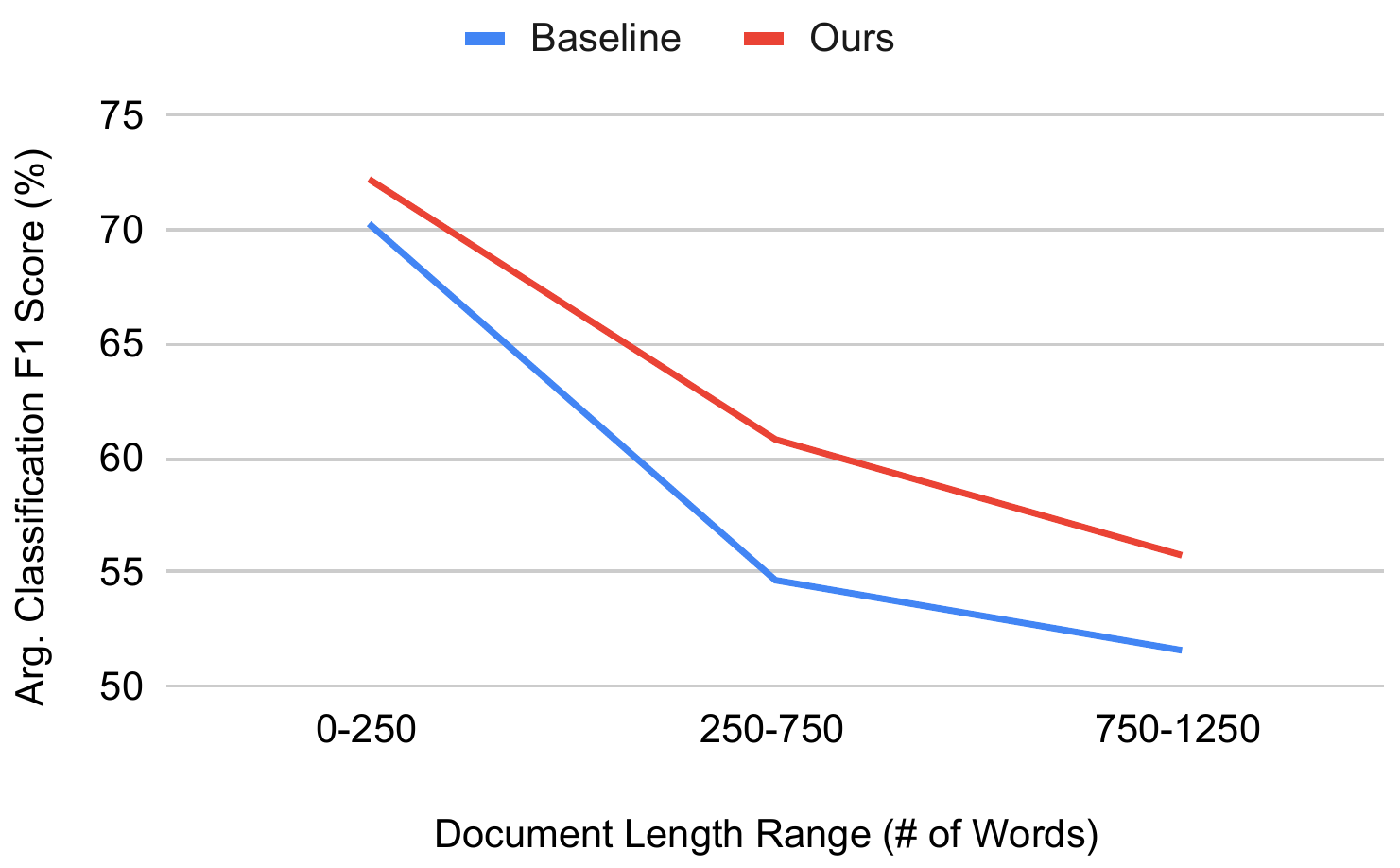}
    \end{subfigure} 
    \hfill \hfill

    \begin{subfigure}[b]{\columnwidth}
    \centering
    \includegraphics[width=\columnwidth]{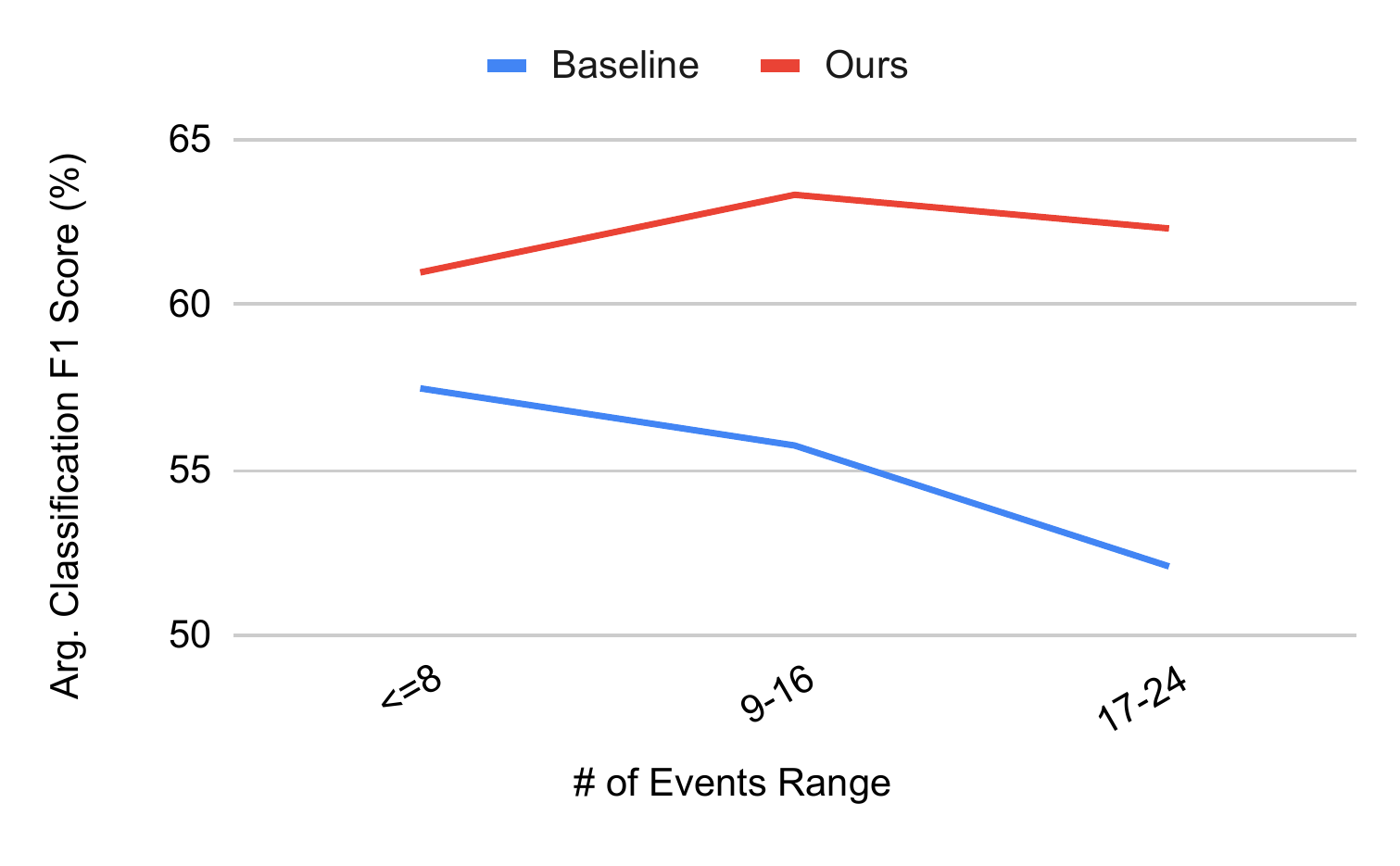}
    \end{subfigure}
    \caption{Effect of doc length and events \# per doc.}
    \label{fig:chart_doclen_evennum}
\end{figure}

\paragraph{Document Length and \# of Events}
In Figure~\ref{fig:chart_doclen_evennum}, we examine how performances change as the document length and the number of events per document grow. First we observe that as the document length grows, challenges grow for both the baseline and our framework (F-1 drops from over 70\% to around 55\%). While our framework maintains a larger advantage when document is longer than 250 words. 
As the number of events per document grows (from <=8 to around 25), our model's performance is not affected much (F-1 all over 60\%). While the baseline system's F-1 score drops to around 50\%.

\begin{table*}[t]
\centering
\resizebox{\textwidth}{!}{
\begin{tabular}{l|l|l|l}
\toprule
& BART-Gen Baseline      & Memory-enhanced Training   & w/ Constrained Decoding  \\ \midrule

Input Doc. 1 & \multicolumn{3}{l}{\begin{tabular}[c]{@{}l@{}}{[}S1{]} ... Accused New York bomber \textcolor{green!80!gray}{Ahmad Khan Rahimi} on Thursday to {\it federal charges} that he set off ...\\ {[}S4{]} ... He spoke only once, when U.S. District Judge Richard Berman asked him to ...\\ {[}S9{]} The confrontation left him with several gunshot {\bf wounds}, delaying the filing of {\it federal charges} ... \end{tabular}}           \\ \midrule

Decoded Seq. 
& \begin{tabular}[c]{@{}l@{}}\textcolor{red}{Richard Berman[{\sc Victim}]} \\ was injured by <arg> ...
\end{tabular} 
& \begin{tabular}[c]{@{}l@{}}\textcolor{green!80!gray}{Ahmad Khan Rahimi[{\sc Victim}]} \\ was injured by <arg> ...
\end{tabular} 
& \begin{tabular}[c]{@{}l@{}}\textcolor{green!80!gray}{Ahmad Khan Rahimi[{\sc Victim}]} \\ was injured by <arg> ...
\end{tabular} \\ \midrule

Input Doc. 2& \multicolumn{3}{l}{\begin{tabular}[c]{@{}l@{}}{[}S1{]} Cuba {\bf sidesteps} Colombia 2019s request to ... \\ {[}S11{]} In November, Colombia asked Cuba to {\bf capture} ELN rebel commander \textcolor{green!80!gray}{Nicolas Rodriguez} \\ and provide information about the presence of other commanders in the Cuban territory. ... \\ {[}S13{]} The Cuban government did not respond publicly to that request or made a statement ...\end{tabular}}    \\ \midrule

Decoded Seq.  
& \begin{tabular}[c]{@{}l@{}}\textcolor{red}{Cuba[{\sc Jailer}]} arrested or jailed \\ \textcolor{green!80!gray}{Nicolas Rodriguez[{\sc Detainee}]} ... \end{tabular} 
& \begin{tabular}[c]{@{}l@{}}\textcolor{red}{Cuba[{\sc Jailer}]} arrested or jailed \\ \textcolor{green!80!gray}{Nicolas Rodriguez[{\sc Detainee}]} ... \end{tabular} 
& \begin{tabular}[c]{@{}l@{}}<arg> arrested or jailed \\ \textcolor{green!80!gray}{Nicolas Rodriguez[{\sc Detainee}]} ... \end{tabular} \\\bottomrule
\end{tabular}
}
\caption{Decoded Seq. (Extracted Arguments) by BART-Gen and Our Models.}
\label{tab:qualitative_eg}
\end{table*}

\paragraph{Qualitative Analysis}

We present a couple of representative examples (Table~\ref{tab:qualitative_eg}). In the first example, for the event triggered by {\bf wounds}, it's hard to find the {\sc Victim} argument ``Ahmad Khan Rahimi'' since it's explicitly mentioned far before the current sentence. But with retrieved additional context, both our framework variants are able to extract the full name correctly. In the second  example, ``Cuba'' was mentioned in two sentences with two events (Impede event triggered by {\bf sidesteps} and Arrest triggered by {\bf capture}). But it only participated in the first event. According to our argument pair constraints -- it's improbable that one entity is both an {\sc Impeder} and a {\sc Jailer}, our framework with constrained decoding conducts reasoning to avoid the wrong extraction.

\begin{table}[t]
\resizebox{\columnwidth}{!}{
\begin{tabular}{l|ccc}
\toprule
        & Missing & Spurious & Misclassified \\\midrule
Head M  & 239 (52.88\%) & 	187 (41.37\%)  & 26 (5.75\%)       \\
Coref M & 213 (52.85\%) & 161 (39.95\%)  & 29 (7.20\%)      \\\bottomrule
\end{tabular}}
\caption{Types of Errors Made by Our Framework.}
\label{tab:error_types}
\end{table}

\paragraph{Error Analysis and Remaining Challenges}

\begin{figure}[t]
\resizebox{0.95\columnwidth}{!}{
\includegraphics{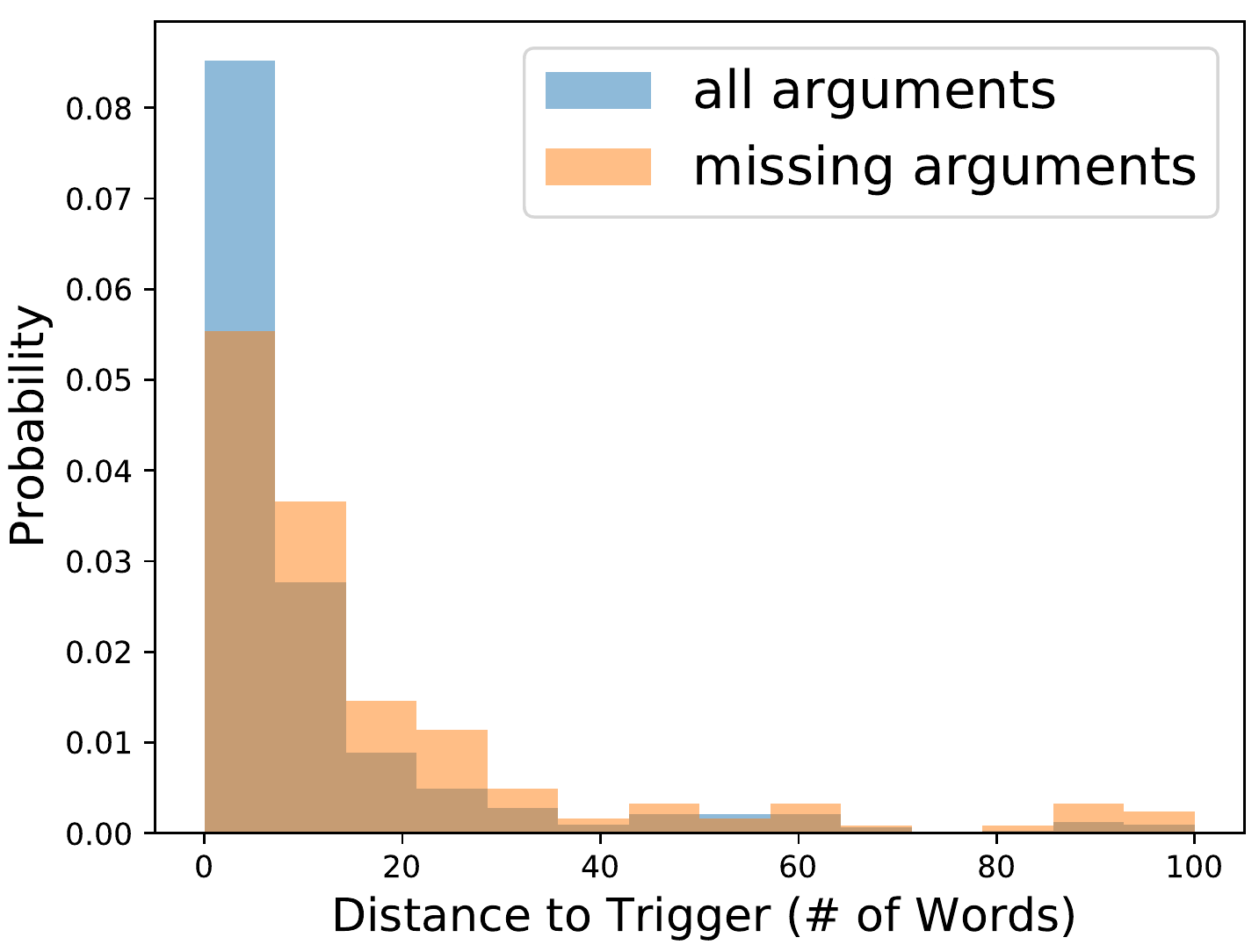}
}
\caption{Distribution of Distance between Informative Arguments and the Gold-standard Triggers.}
\label{fig:arg_dist}
\end{figure}

Table~\ref{tab:error_types} categorizes types of argument extraction errors made by our best model. The majority of errors is from missing arguments and only around 7\% of cases are caused by incorrectly-assigned argument roles (e.g., a  {\sc Place} argument is mistakenly labeled as a {\sc Target} argument).
Interestingly, from Figure~\ref{fig:arg_dist}'s distribution, we see that as compared to the distance of gold-standard informative arguments to the trigger (avg. 80.41 words), the missing arguments are far away (avg. 136.39 words) -- showing the hardness of extracting distant arguments as compared to local arguments.

Finally we examine deeper the example predictions and categorize reasons for errors into the following types:
%
    (1) {\it Challenge to obtain an accurate boundary of the argument span.} In the example excerpt ``On Sunday, a suicide bombing in the southeastern province of [Logar] left eight ...'', our model extracts ``southeastern province'' as {\sc Place}.
    Similarly in ``... were transported to [Kabul] city..'', our model extracts ``city'' as {\sc Destination}. In both cases the model gets no credit. To mitigate this problem, models should be able to identify certain noun phrase boundaries with external knowledge.
    Plus,  the improvement of data annotation and evaluation is also needed -- the model should get certain credit though the span does not overlap but related to the gold argument.
    (2) {\it Long distance dependency and deeper context understanding.} In news, most of the contents are written by the author while certain content is cited from participants. While models usually do not distinguish the difference and consider the big stance difference. In the excerpt ``Bill Richard, whose son, Martin, was the youngest person killed in the bombing, said Tsarnaev could have backed out ... Instead, \uwave{Richard said, he chose hate. he chose destruction. He chose death. ...}'', the full name of the informative argument (``D. Tsarnaev'') was mentioned at the very beginning of the document. Although our model can leverage previously decoded events, it is not able to fully understand the speaker's point of view and misses the full {\sc Killer} argument span.

\section{Related Work}

\paragraph{Event Knowledge}
There has been work on acquiring event-event knowledge/subevent knowledge with heuristic-based rules or crowdsourcing-based methods.
\newcite{Sap-etal-2019-atomic} propose to use crowdsourcing for obtaining \textit{if-then} relations between events.
\newcite{bosselut-etal-2019-comet} use generative language models to generate new event knowledge based on crowdsourced triples.
\newcite{yao-etal-2020-weakly} propose a weakly-supervised approach to extract sub-event relation tuples from the text.
In our work, we focus on harvesting knowledge-based event argument pair constraints from the predefined ontology with training data co-occurrence statistics. Plus, the work above on knowledge acquisition has not investigated explicitly encoding the knowledge/constraints for improving the performance of models of document-level event extraction related tasks.

\paragraph{Document-level Event Extraction}
Event extraction has been mainly studied under the document-level setting (the template filling tasks from the MUC conferences~\cite{grishman-sundheim-1996-message}) and the sentence-level setting (using the ACE data~\cite{doddington-etal-2004-automatic} and BioNLP shared tasks~\cite{kim-etal-2009-overview}). 
In this paper, we focus on the document-level event argument extraction task which is a less-explored and challenging topic~\cite{du-etal-2021-template, li-etal-2021-document}. To support the progress for the problem, \newcite{ebner-etal-2020-multi} built RAMS dataset, and it contains annotations for cross-sentence arguments but for each document it contains only one event. Later \newcite{li-etal-2021-document} built the benchmark {\sc WikiEvents} with complete event annotations for each document.
Regarding the methodology, neural text generation-based models have been proved to be superior at this document-level task~\cite{huang-etal-2021-document, du-etal-2021-template, li-etal-2021-document}. But they are still limited by the maximum length context issue and mainly focus on modeling one event at a time. \newcite{yang-mitchell-2016-joint} proposed a joint extraction approach that models cross-event dependencies -- but it's restricted to events co-occurring within a sentence and only does trigger typing. In our framework, utilizing the memory store can help better capture global context and avoid  the document length constraint.
Apart from event extraction, in the future, it's worth investigating how to leverage the global memory idea for other document-level IE problems like ($N$-ary) relation extraction~\cite{quirk-poon-2017-distant, yao-etal-2019-docred}.

\section{Conclusions and Future Work}
In this work, we examined the effect of global document-level ``memory'' on informative event argument extraction. In the new framework, we propose to leverage the previously extracted  events as additional context to help the model learn the dependency across events. At test time, we propose to use a dynamic decoding process to help the model satisfy global knowledge-based argument constraints. Experiments demonstrate that our approach achieves substantial improvements over prior methods and has a larger advantage when document length and events number increase.
For future work, we plan to investigate how to extend our method to multi-document event extraction cases.

\section*{Acknowledgement}
We thank the anonymous reviewers helpful suggestions. 
This research is based upon work supported by U.S. DARPA KAIROS Program No. FA8750-19-2-1004, U.S. DARPA AIDA Program No. FA8750-18-2-0014 and LORELEI Program No. HR0011-15-C-0115. The views and conclusions contained herein are those of the authors and should not be interpreted as necessarily representing the official policies, either expressed or implied, of DARPA, or the U.S. Government. The U.S. Government is authorized to reproduce and distribute reprints for governmental purposes notwithstanding any copyright annotation therein.

\bibliography{custom}

\begin{thebibliography}{30}
\expandafter\ifx\csname natexlab\endcsname\relax\def\natexlab#1{#1}\fi

\bibitem[{Berg-Kirkpatrick et~al.(2012)Berg-Kirkpatrick, Burkett, and
  Klein}]{berg-etal-2012-empirical}
Taylor Berg-Kirkpatrick, David Burkett, and Dan Klein. 2012.
\newblock \href {https://aclanthology.org/D12-1091} {An empirical investigation
  of statistical significance in {NLP}}.
\newblock In \emph{Proceedings of the 2012 Joint Conference on Empirical
  Methods in Natural Language Processing and Computational Natural Language
  Learning}, pages 995--1005, Jeju Island, Korea. Association for Computational
  Linguistics.

\bibitem[{Bosselut et~al.(2019)Bosselut, Rashkin, Sap, Malaviya, Celikyilmaz,
  and Choi}]{bosselut-etal-2019-comet}
Antoine Bosselut, Hannah Rashkin, Maarten Sap, Chaitanya Malaviya, Asli
  Celikyilmaz, and Yejin Choi. 2019.
\newblock \href {https://doi.org/10.18653/v1/P19-1470} {{COMET}: Commonsense
  transformers for automatic knowledge graph construction}.
\newblock In \emph{Proceedings of the 57th Annual Meeting of the Association
  for Computational Linguistics}, pages 4762--4779, Florence, Italy.
  Association for Computational Linguistics.

\bibitem[{Devlin et~al.(2019)Devlin, Chang, Lee, and
  Toutanova}]{devlin-etal-2019-bert}
Jacob Devlin, Ming-Wei Chang, Kenton Lee, and Kristina Toutanova. 2019.
\newblock \href {https://doi.org/10.18653/v1/N19-1423} {{BERT}: Pre-training of
  deep bidirectional transformers for language understanding}.
\newblock In \emph{Proceedings of the 2019 Conference of the North {A}merican
  Chapter of the Association for Computational Linguistics: Human Language
  Technologies, Volume 1 (Long and Short Papers)}, pages 4171--4186,
  Minneapolis, Minnesota. Association for Computational Linguistics.

\bibitem[{Doddington et~al.(2004)Doddington, Mitchell, Przybocki, Ramshaw,
  Strassel, and Weischedel}]{doddington-etal-2004-automatic}
George Doddington, Alexis Mitchell, Mark Przybocki, Lance Ramshaw, Stephanie
  Strassel, and Ralph Weischedel. 2004.
\newblock \href {http://www.lrec-conf.org/proceedings/lrec2004/pdf/5.pdf} {The
  automatic content extraction ({ACE}) program {--} tasks, data, and
  evaluation}.
\newblock In \emph{Proceedings of the Fourth International Conference on
  Language Resources and Evaluation ({LREC}{'}04)}, Lisbon, Portugal. European
  Language Resources Association (ELRA).

\bibitem[{Du(2021)}]{du-thesis-2021}
Xinya Du. 2021.
\newblock \href
  {https://www.proquest.com/dissertations-theses/towards-more-intelligent-extraction-information/docview/2581838694/se-2?accountid=10267}
  {\emph{Towards More Intelligent Extraction of Information from Documents}}.
\newblock Ph.D. thesis, Cornell University.
\newblock Copyright - Database copyright ProQuest LLC; ProQuest does not claim
  copyright in the individual underlying works.

\bibitem[{Du and Cardie(2020)}]{du-cardie-2020-event}
Xinya Du and Claire Cardie. 2020.
\newblock \href {https://doi.org/10.18653/v1/2020.emnlp-main.49} {Event
  extraction by answering (almost) natural questions}.
\newblock In \emph{Proceedings of the 2020 Conference on Empirical Methods in
  Natural Language Processing (EMNLP)}, pages 671--683, Online. Association for
  Computational Linguistics.

\bibitem[{Du et~al.(2021)Du, Rush, and Cardie}]{du-etal-2021-template}
Xinya Du, Alexander Rush, and Claire Cardie. 2021.
\newblock \href {https://doi.org/10.18653/v1/2021.naacl-main.70} {Template
  filling with generative transformers}.
\newblock In \emph{Proceedings of the 2021 Conference of the North American
  Chapter of the Association for Computational Linguistics: Human Language
  Technologies}, pages 909--914, Online. Association for Computational
  Linguistics.

\bibitem[{Ebner et~al.(2020)Ebner, Xia, Culkin, Rawlins, and
  Van~Durme}]{ebner-etal-2020-multi}
Seth Ebner, Patrick Xia, Ryan Culkin, Kyle Rawlins, and Benjamin Van~Durme.
  2020.
\newblock \href {https://doi.org/10.18653/v1/2020.acl-main.718} {Multi-sentence
  argument linking}.
\newblock In \emph{Proceedings of the 58th Annual Meeting of the Association
  for Computational Linguistics}, pages 8057--8077, Online. Association for
  Computational Linguistics.

\bibitem[{Grishman(2019)}]{grishman_2019}
Ralph Grishman. 2019.
\newblock \href {https://doi.org/10.1017/S1351324919000512} {Twenty-five years
  of information extraction}.
\newblock \emph{Natural Language Engineering}, 25(6):677–692.

\bibitem[{Grishman and Sundheim(1996)}]{grishman-sundheim-1996-message}
Ralph Grishman and Beth Sundheim. 1996.
\newblock \href {https://aclanthology.org/C96-1079} {{M}essage {U}nderstanding
  {C}onference- 6: A brief history}.
\newblock In \emph{{COLING} 1996 Volume 1: The 16th International Conference on
  Computational Linguistics}.

\bibitem[{Huang et~al.(2021)Huang, Tang, and Peng}]{huang-etal-2021-document}
Kung-Hsiang Huang, Sam Tang, and Nanyun Peng. 2021.
\newblock \href {https://aclanthology.org/2021.emnlp-main.426} {Document-level
  entity-based extraction as template generation}.
\newblock In \emph{Proceedings of the 2021 Conference on Empirical Methods in
  Natural Language Processing}, pages 5257--5269, Online and Punta Cana,
  Dominican Republic. Association for Computational Linguistics.

\bibitem[{Huang and Riloff(2012)}]{huang2012modeling}
Ruihong Huang and Ellen Riloff. 2012.
\newblock Modeling textual cohesion for event extraction.
\newblock In \emph{Proceedings of the AAAI Conference on Artificial
  Intelligence}.

\bibitem[{Ji and Grishman(2008{\natexlab{a}})}]{ji-grishman-2008-refining}
Heng Ji and Ralph Grishman. 2008{\natexlab{a}}.
\newblock \href {https://aclanthology.org/P08-1030} {Refining event extraction
  through cross-document inference}.
\newblock In \emph{Proceedings of ACL-08: HLT}, pages 254--262, Columbus, Ohio.
  Association for Computational Linguistics.

\bibitem[{Ji and Grishman(2008{\natexlab{b}})}]{ji2008refining}
Heng Ji and Ralph Grishman. 2008{\natexlab{b}}.
\newblock Refining event extraction through unsupervised cross-document
  inference.
\newblock In \emph{In Proceedings of the Annual Meeting of the Association of
  Computational Linguistics (ACL 2008). Ohio, USA}.

\bibitem[{Khandelwal et~al.(2018)Khandelwal, He, Qi, and
  Jurafsky}]{khandelwal-etal-2018-sharp}
Urvashi Khandelwal, He~He, Peng Qi, and Dan Jurafsky. 2018.
\newblock \href {https://doi.org/10.18653/v1/P18-1027} {Sharp nearby, fuzzy far
  away: How neural language models use context}.
\newblock In \emph{Proceedings of the 56th Annual Meeting of the Association
  for Computational Linguistics (Volume 1: Long Papers)}, pages 284--294,
  Melbourne, Australia. Association for Computational Linguistics.

\bibitem[{Kim et~al.(2009)Kim, Ohta, Pyysalo, Kano, and
  Tsujii}]{kim-etal-2009-overview}
Jin-Dong Kim, Tomoko Ohta, Sampo Pyysalo, Yoshinobu Kano, and Jun{'}ichi
  Tsujii. 2009.
\newblock \href {https://aclanthology.org/W09-1401} {Overview of
  {B}io{NLP}{'}09 shared task on event extraction}.
\newblock In \emph{Proceedings of the {B}io{NLP} 2009 Workshop Companion Volume
  for Shared Task}, pages 1--9, Boulder, Colorado. Association for
  Computational Linguistics.

\bibitem[{Lafferty et~al.(2001)Lafferty, McCallum, and
  Pereira}]{DBLP:conf/icml/LaffertyMP01}
John~D. Lafferty, Andrew McCallum, and Fernando C.~N. Pereira. 2001.
\newblock Conditional random fields: Probabilistic models for segmenting and
  labeling sequence data.
\newblock In \emph{Proceedings of the Eighteenth International Conference on
  Machine Learning {(ICML} 2001), Williams College, Williamstown, MA, USA, June
  28 - July 1, 2001}, pages 282--289. Morgan Kaufmann.

\bibitem[{Lewis et~al.(2020)Lewis, Liu, Goyal, Ghazvininejad, Mohamed, Levy,
  Stoyanov, and Zettlemoyer}]{lewis-etal-2020-bart}
Mike Lewis, Yinhan Liu, Naman Goyal, Marjan Ghazvininejad, Abdelrahman Mohamed,
  Omer Levy, Veselin Stoyanov, and Luke Zettlemoyer. 2020.
\newblock \href {https://doi.org/10.18653/v1/2020.acl-main.703} {{BART}:
  Denoising sequence-to-sequence pre-training for natural language generation,
  translation, and comprehension}.
\newblock In \emph{Proceedings of the 58th Annual Meeting of the Association
  for Computational Linguistics}, pages 7871--7880, Online. Association for
  Computational Linguistics.

\bibitem[{Li et~al.(2013)Li, Ji, and Huang}]{li-etal-2013-joint}
Qi~Li, Heng Ji, and Liang Huang. 2013.
\newblock \href {https://aclanthology.org/P13-1008} {Joint event extraction via
  structured prediction with global features}.
\newblock In \emph{Proceedings of the 51st Annual Meeting of the Association
  for Computational Linguistics (Volume 1: Long Papers)}, pages 73--82, Sofia,
  Bulgaria. Association for Computational Linguistics.

\bibitem[{Li et~al.(2021)Li, Ji, and Han}]{li-etal-2021-document}
Sha Li, Heng Ji, and Jiawei Han. 2021.
\newblock \href {https://doi.org/10.18653/v1/2021.naacl-main.69}
  {Document-level event argument extraction by conditional generation}.
\newblock In \emph{Proceedings of the 2021 Conference of the North American
  Chapter of the Association for Computational Linguistics: Human Language
  Technologies}, pages 894--908, Online. Association for Computational
  Linguistics.

\bibitem[{Lin et~al.(2020)Lin, Ji, Huang, and Wu}]{LinACL2020}
Ying Lin, Heng Ji, Fei Huang, and Lingfei Wu. 2020.
\newblock A joint end-to-end neural model for information extraction with
  global features.
\newblock In \emph{Proc. The 58th Annual Meeting of the Association for
  Computational Linguistics (ACL2020)}.

\bibitem[{Quirk and Poon(2017)}]{quirk-poon-2017-distant}
Chris Quirk and Hoifung Poon. 2017.
\newblock \href {https://aclanthology.org/E17-1110} {Distant supervision for
  relation extraction beyond the sentence boundary}.
\newblock In \emph{Proceedings of the 15th Conference of the {E}uropean Chapter
  of the Association for Computational Linguistics: Volume 1, Long Papers},
  pages 1171--1182, Valencia, Spain. Association for Computational Linguistics.

\bibitem[{Reimers and Gurevych(2019)}]{reimers-gurevych-2019-sentence}
Nils Reimers and Iryna Gurevych. 2019.
\newblock \href {https://doi.org/10.18653/v1/D19-1410} {Sentence-{BERT}:
  Sentence embeddings using {S}iamese {BERT}-networks}.
\newblock In \emph{Proceedings of the 2019 Conference on Empirical Methods in
  Natural Language Processing and the 9th International Joint Conference on
  Natural Language Processing (EMNLP-IJCNLP)}, pages 3982--3992, Hong Kong,
  China. Association for Computational Linguistics.

\bibitem[{Sap et~al.(2019)Sap, Bras, Allaway, Bhagavatula, Lourie, Rashkin,
  Roof, Smith, and Choi}]{Sap-etal-2019-atomic}
Maarten Sap, Ronan~Le Bras, Emily Allaway, Chandra Bhagavatula, Nicholas
  Lourie, Hannah Rashkin, Brendan Roof, Noah~A. Smith, and Yejin Choi. 2019.
\newblock \href {https://doi.org/10.1609/aaai.v33i01.33013027} {{ATOMIC:} an
  atlas of machine commonsense for if-then reasoning}.
\newblock In \emph{The Thirty-Third {AAAI} Conference on Artificial
  Intelligence, {AAAI} 2019, The Thirty-First Innovative Applications of
  Artificial Intelligence Conference, {IAAI} 2019, The Ninth {AAAI} Symposium
  on Educational Advances in Artificial Intelligence, {EAAI} 2019, Honolulu,
  Hawaii, USA, January 27 - February 1, 2019}, pages 3027--3035. {AAAI} Press.

\bibitem[{Shi and Lin(2019)}]{shi-lin-2019-BERTCRF}
Peng Shi and Jimmy Lin. 2019.
\newblock \href {http://arxiv.org/abs/1904.05255} {Simple {BERT} models for
  relation extraction and semantic role labeling}.
\newblock \emph{CoRR}, abs/1904.05255.

\bibitem[{Sundheim(1992)}]{sundheim-1992-overview}
Beth~M. Sundheim. 1992.
\newblock \href {https://aclanthology.org/M92-1001} {Overview of the fourth
  {M}essage {U}nderstanding {E}valuation and {C}onference}.
\newblock In \emph{{F}ourth {M}essage {U}understanding {C}onference ({MUC}-4):
  Proceedings of a Conference Held in {M}c{L}ean, {V}irginia, {J}une 16-18,
  1992}.

\bibitem[{Wadden et~al.(2019)Wadden, Wennberg, Luan, and
  Hajishirzi}]{wadden-etal-2019-entity}
David Wadden, Ulme Wennberg, Yi~Luan, and Hannaneh Hajishirzi. 2019.
\newblock \href {https://doi.org/10.18653/v1/D19-1585} {Entity, relation, and
  event extraction with contextualized span representations}.
\newblock In \emph{Proceedings of the 2019 Conference on Empirical Methods in
  Natural Language Processing and the 9th International Joint Conference on
  Natural Language Processing (EMNLP-IJCNLP)}, pages 5784--5789, Hong Kong,
  China. Association for Computational Linguistics.

\bibitem[{Yang and Mitchell(2016)}]{yang-mitchell-2016-joint}
Bishan Yang and Tom~M. Mitchell. 2016.
\newblock \href {https://doi.org/10.18653/v1/N16-1033} {Joint extraction of
  events and entities within a document context}.
\newblock In \emph{Proceedings of the 2016 Conference of the North {A}merican
  Chapter of the Association for Computational Linguistics: Human Language
  Technologies}, pages 289--299, San Diego, California. Association for
  Computational Linguistics.

\bibitem[{Yao et~al.(2020)Yao, Dai, Ramaswamy, Min, and
  Huang}]{yao-etal-2020-weakly}
Wenlin Yao, Zeyu Dai, Maitreyi Ramaswamy, Bonan Min, and Ruihong Huang. 2020.
\newblock \href {https://doi.org/10.18653/v1/2020.emnlp-main.430} {Weakly
  {S}upervised {S}ubevent {K}nowledge {A}cquisition}.
\newblock In \emph{Proceedings of the 2020 Conference on Empirical Methods in
  Natural Language Processing (EMNLP)}, pages 5345--5356, Online. Association
  for Computational Linguistics.

\bibitem[{Yao et~al.(2019)Yao, Ye, Li, Han, Lin, Liu, Liu, Huang, Zhou, and
  Sun}]{yao-etal-2019-docred}
Yuan Yao, Deming Ye, Peng Li, Xu~Han, Yankai Lin, Zhenghao Liu, Zhiyuan Liu,
  Lixin Huang, Jie Zhou, and Maosong Sun. 2019.
\newblock \href {https://doi.org/10.18653/v1/P19-1074} {{D}oc{RED}: A
  large-scale document-level relation extraction dataset}.
\newblock In \emph{Proceedings of the 57th Annual Meeting of the Association
  for Computational Linguistics}, pages 764--777, Florence, Italy. Association
  for Computational Linguistics.

\end{thebibliography}
\bibliographystyle{acl_natbib}

\clearpage

\appendix

\clearpage

\section{Examples of Argument Pairs}

We list a couple of improbable argument pairs from the ``checklist''.

\begin{table}[h]
\resizebox{\textwidth}{!}{
\begin{tabular}{ll|ll}
\toprule
\multicolumn{2}{c|}{Argument 1}                       & \multicolumn{2}{c}{Argument 2}                         \\ \midrule
Justice.Sentence.Unspecified            & JudgeCourt & Life.Die.Unspecified                    & Victim       \\
Justice.Sentence.Unspecified            & Defendant  & Life.Die.Unspecified                    & Victim       \\
Control.ImpedeInterfereWith.Unspecified & Impeder    & Justice.ArrestJailDetain.Unspecified    & Jailer       \\
Contact.RequestCommand.Unspecified      & Recipient  & Justice.ArrestJailDetain.Unspecified    & Jailer       \\
Life.Injure.Unspecified                 & Injurer    & Transaction.ExchangeBuySell.Unspecified & Giver        \\
Justice.TrialHearing.Unspecified        & Defendant  & Transaction.ExchangeBuySell.Unspecified & Giver        \\
Justice.TrialHearing.Unspecified        & Defendant  & Transaction.ExchangeBuySell.Unspecified & Recipient    \\
Conflict.Attack.DetonateExplode         & Attacker   & Contact.Contact.Broadcast               & Communicator \\
Conflict.Attack.Unspecified             & Attacker   & Contact.Contact.Broadcast               & Communicator \\
Conflict.Attack.DetonateExplode         & Attacker   & Contact.ThreatenCoerce.Unspecified      & Communicator \\
Conflict.Attack.Unspecified             & Attacker   & Contact.ThreatenCoerce.Unspecified      & Communicator \\ \bottomrule
\end{tabular}}
\end{table}

\section{Hyperparameters used in The Experiments}

\begin{table}[h]
\begin{tabular}{l|l}
\toprule
train batch size        & 2    \\
eval batch size         & 1    \\
learning rate           & 3e-5 \\
accumulate grad batches & 4    \\
training epoches        & 5    \\
warmup steps            & 0    \\
weight decay            & 0    \\
\# gpus                 & 1    \\ \bottomrule
\end{tabular}
\caption{Hyperparameters.}
\end{table}


\end{document}